\documentclass[letterpaper, 10 pt, conference]{ieeeconf}   
\IEEEoverridecommandlockouts                              
  
\usepackage{enumerate}
\usepackage{graphics} 
\usepackage{epsfig} 
\usepackage{amsmath} 
\usepackage{amssymb}  
\usepackage{array}
\usepackage{subcaption}
\usepackage{leftidx}
\usepackage{hyperref}
\usepackage[tight]{units}
\usepackage{color}

\renewcommand{\theequation}{CT\arabic{equation}}

\newcommand{\co}{$\mathcal{C}$}
\newcommand{\cfree}{$\mathcal{C}_{free}({\mathcal{O}})$}

\newcommand{\cso}{$\mathcal{C}(\theta_{s})$}
\newcommand{\csfree}{$\mathcal{C}_{free}(\mathcal{O},\theta_s)$}

\title{\LARGE \bf A Convex-Combinatorial Model for Planar Caging}

\author{\hspace{-0pt} Bernardo Aceituno-Cabezas$^{1}$, Hongkai Dai$^{2}$, and Alberto Rodriguez$^{1}$  \\ \\
$^{1}$Department of Mechanical Engineering --- Massachusetts Institute of Technology\\
$^{2}$Toyota Research Institute\\ 
{\tt \small <aceituno,albertor>@mit.edu}, {\tt \small hongkai.dai@tri.global}\\
}

\begin{document}

\maketitle
\begin{abstract}
Caging is a promising tool which allows a robot to manipulate an object without directly reasoning about the contact dynamics involved. Furthermore, caging also provides useful guarantees in terms of robustness to uncertainty, and often serves as a way-point to a grasp. However, caging is traditionally difficult to integrate as part of larger manipulation frameworks, where caging is not the goal but an intermediate condition.
In this paper, we develop a convex-combinatorial model to characterize caging from an optimization perspective. More specifically, we derive a set of sufficient constraints to enclose the configuration of the object in a compact-connected component of its free-space. The convex-combinatorial nature of this approach provides guarantees on optimality and convergence, and its optimization nature makes it versatile for further applications on robot manipulation tasks. To the best of our knowledge, this is the first optimization-based approach to formulate the caging condition.
\end{abstract}

\section{Introduction}

A cage is an arrangement of obstacles that bounds the mobility of an object. 
The connection of cages to invariant regions and robustness has attracted the attention of the manipulation community for a long time.
A cage can be used as a waypoint to a grasp~\cite{wan2012grasping,rodriguez2012caging}, providing a guarantee that the object will not escape in the process.
A cage can also be used to manipulate without rigid immobilization, alleviating common issues from jamming, wedging, and general over-constrained interactions (e.g. turning the handle of a door).

The reality, however, is that the practical applications of existing caging algorithms have been limited.
In this work we propose to rethink the conventional 
topologic/geometric approach to characterize caging---focused on developing complete algorithms to analyze the configuration space of an object surrounded by a set of obstacles. Here, we develop an approach aimed at the synthesis of manipulation strategies that can incorporate and exploit the caging condition, the same way that we normally use grasping for immobilization. 

\begin{figure}[t]
    \centering
    \includegraphics[width=0.9\linewidth]{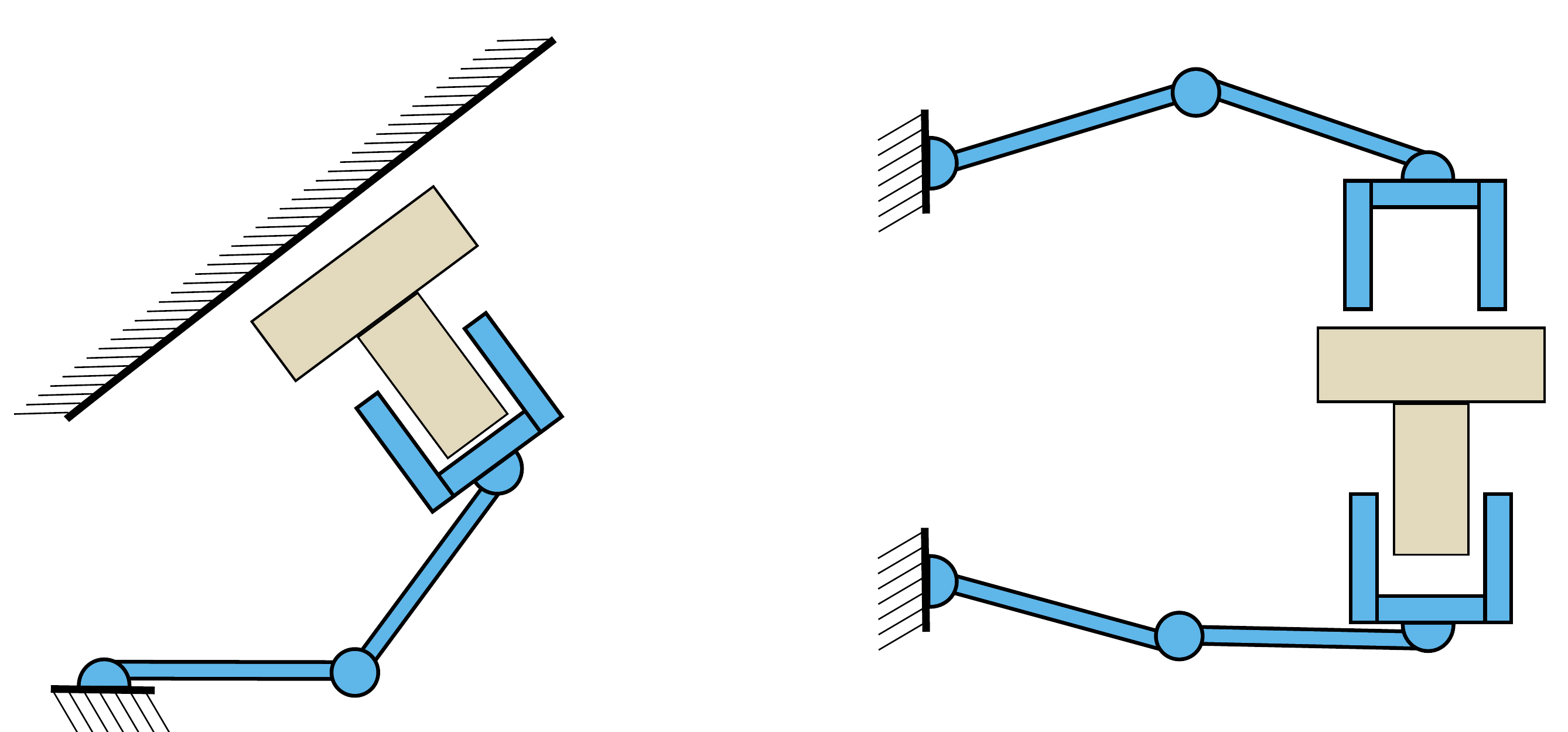}
    \caption{\small Examples of caging in the context of robot manipulation: trapping against a wall and caging under kinematic constraints.}
    \label{fig:my_label}
    \vspace{-12pt}
\end{figure}

Figure \ref{fig:my_label} describes the motivation and the long-term goal of this project: How do we synthesize a manipulation plan to cage an object while exploiting environment constraints and respecting the kinematic constraints of the robot?
The work in this paper is a first step in that direction. We present a reformulation of the caging condition as a set of mixed-integer convex constraints that provide:
%
%
%
\begin{itemize}
    \item \textbf{Versatility} to incorporate the caging condition in the context of a larger manipulation planning framework.
    \item \textbf{Guarantees} of the optimization framework that steams from the convex-combinatorial nature of the model. If a cage exists within the set of conditions described by the mixed-integer problem, the optimization algorithm will find it. If it does not exist, it will report so. 
\end{itemize}

These come at the expense of exponential bounds on computation time, and resolution completeness of the algorithm. In particular, this paper focuses on caging polygonal objects---described as the union of convex polygons---with a manipulator described by an arbitrary number of point fingers in the plane. To do this, the proposed approach provides a set of sufficient conditions to cage such object. The main contributions of this paper are:
\begin{itemize}
    \item \textbf{Cage optimization} algorithm based on the proposed convex-combinatorial caging model. This can be computed efficiently with off-the-shelf mixed-integer solvers yielding global convergence guarantees.
    \item \textbf{Validation} of the proposed cage synthesis algorithm on random planar polygons.
    \item \textbf{Application} of the caging algorithm to find cages that either exploit the environment (walls) or take into account kinematic limits in finger motions.
\end{itemize}


In Section VII, we discuss the possibility of using the approach to synthesize cage-reach-grasp motions with formal guarantees of convergence, and its application to design gripper kinematics and shapes defined by a set of points.

\subsection{Related Work}

Algorithms for caging have been studied since Rimon and Blake \cite{rimon1999caging} introduced this notion to the robotics community. The first caging algorithms proposed were focused on characterizing the set of point-finger configurations capable of caging a polygon \cite{rimon1999caging,sudsang2000new,vahedi2008caging}. Since then, cage-finding algorithms have continuously improved. Some works have formulated the problem in contact0space, providing better computational complexity \cite{allen2015two,bunis2018equilateral}. Some other works have shown how tools from computational topology can be used to find and verify cages on 2D and 3D objects  \cite{varava2016caging,mahler2016energy}. Furthermore, the work of Varava and Carvalho et. al. \cite{varava2017caging} showed how sampling-based methods could efficiently find cages of 2D and 3D objects with arbitrary manipulator geometries. In similar spirit to our work, Pererira et. al. and Varava et. al. \cite{pereira2004decentralized,varava2017herding} showed the application of caging as a tool for multi-robot transportation of polygonal objects. 

\vspace{6pt}

In comparison to prior work, our model is less computationally efficient and more limited, as the conditions derived are only sufficient to guarantee a cage. However, this approach offers significantly more flexibility, as it can be easily included as constraints within trajectory or shape optimization, while still characterizing a large set of cages.

\section{Preliminaries}

In this section, we introduce the notation that we will use through this paper and define the idea of caging.

\subsection{Notation}

Given an object $\mathcal{O}$ on a workspace $\mathcal{W}$, we will denote its \textit{Configuration Space} \cite{lozano1983spatial} as $\mathcal{C} \subseteq SO(2)$. We refer to a plane of $\mathcal{C}$ with fixed orientation component $\theta_s$ as a $\mathcal{C}-$slice, denoted $\mathcal{C}(\theta_s)$. We refer to an arrangement of point fingers as the manipulator $\mathcal{M}$. We assume $\mathcal{M}$ has $N$ point fingers with positions $\mathcal{M} = \lbrace \boldsymbol{p}_1, \hdots, \boldsymbol{p}_N \rbrace \in \mathcal{W}^N$. \vspace{6pt}

At each $\mathcal{C}-$slice, we refer to the set of configurations where the object penetrates a finger as $\mathcal{C}$-obstacles. Then, the free-space of the object $\mathcal{C}_{free}(\mathcal{O})$ corresponds to the space $\mathcal{C}$ not intersecting any of the $\mathcal{C}$-obstacles, namely the set of object configurations that does not penetrate any of the fingers.

\subsection{Caging}

The caging problem, based on the original formulation by Kuperberg \cite{kuperberg1990problems}, can be stated as:\vspace{6pt}

\textit{For a planar object $\mathcal{O}$ at configuration $q = [q_x, q_y, q_{\theta}]^T$ and a manipulator $\mathcal{M}$, find a configuration of $\mathcal{M}$ such that  $q$ lies in a compact connected component of $\mathcal{C}_{free}(\mathcal{O})$, denoted as $\mathcal{C}^{compact}_{free}(\mathcal{O})$.}\vspace{6pt}

This formulation of the problem, based on topology, is equivalent to the more traditional geometric condition that there exists no continuous path that will drive the object arbitrarily far from the manipulator, as illustrated in Fig. \ref{fig:limit}.
 
\section{Approach Overview}

In this section, we provide an overview on the problem of caging a planar object and describe the conditions to characterize it.

\subsection{Conditions for Caging}

The set of compact-connected components in free-space can be too general to explore. Here, we introduce a set of conditions that characterize a rich set of cages, where the component $\mathcal{C}^{compact}_{free}(\mathcal{O})$ has a single local maxima and minima over the orientation axis. For this, let us first define the idea of limit orientations:\vspace{6pt}

\textbf{Definition 1 (Limit Orientation)}: {Given a compact-connected component $\mathcal{A} \in \mathcal{C}$, its limit orientations $\theta_U, \theta_L$ can be defined as $\theta_U = \sup_{\theta \in \mathcal{A}} \ \theta$ and $\theta_L = \inf_{\theta \in \mathcal{A}} \ \theta$.}\vspace{6pt}

\begin{figure}[t]
		\centering		
        \includegraphics[width=0.99\linewidth]{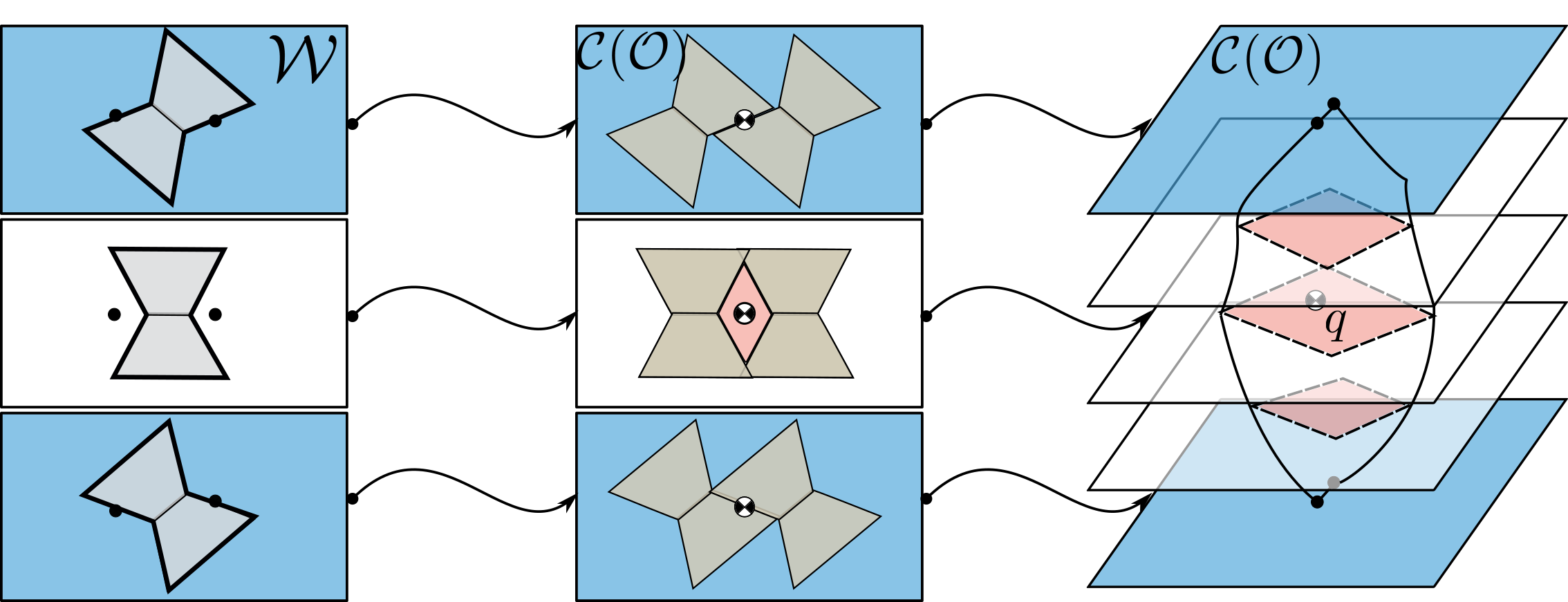}
		\caption{\small Overview of the conditions for caging. Each plane shows a $\mathcal{C}-$slice during a cage between two limit orientations (blue) with a compact-connected component of free-space (red). Note that in the \textit{limit orientations} the object is constrained to a line segment of translational motion by the $\mathcal{C}-$obstacles. Also, the object is only caged if the component remains compact and connected between slices.}
		\vspace{-12pt}
		\label{fig:limit}
\end{figure}

Then, the following conditions generate any component $\mathcal{C}_{free}^{compact}(\mathcal{O})$ with at most two global limit orientations:

\begin{enumerate}
    \item The component is bounded in the orientation axis by two limit orientations, otherwise it infinitely repeats along such axis with period $2 \pi$.
    \item In all $\mathcal{C}-$slices, between two limit orientations when these exist, there is a loop of $\mathcal{C}-$obstacles enclosing a segment of free-space. Such loop encloses $q$ at the slice with $\theta_s = q_\theta$ (as illustrated in the middle column of Fig. \ref{fig:limit}). These loops must be connected, enclosing a component of free-space in between adjacent slices.  
    \item  At the $\mathcal{C}-$slice of the limit orientations, if these exist, the free-space component enclosed by the loop has zero area. Thus, getting reduced to a line segment or a singleton.
\end{enumerate}

\begin{figure}[b]
    \centering
    \vspace{-12pt}
    \includegraphics[width=0.9\linewidth]{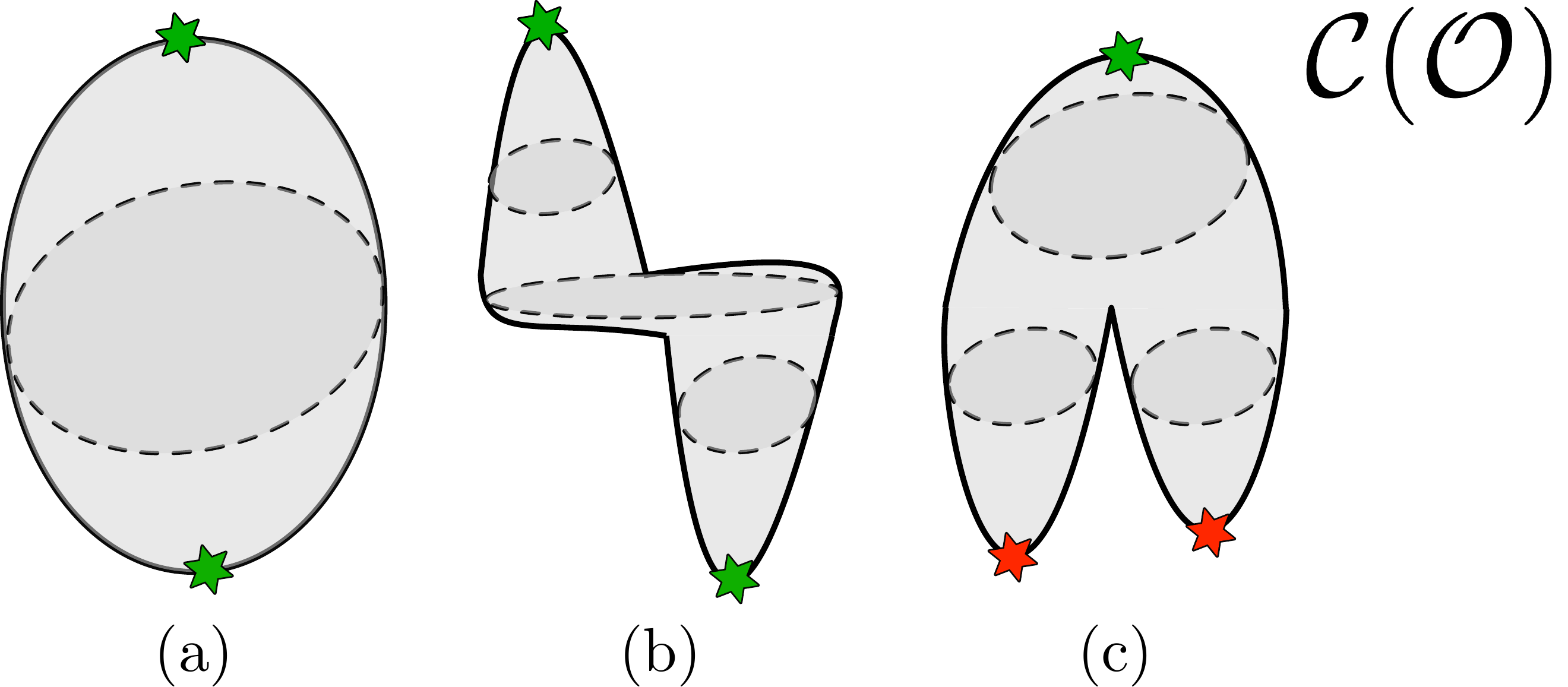}
    \caption{\small Different types of compact connected components. The conditions described in this paper can fully describe components (a) and (b), as both of these components have a pair of limit orientations where the component opens and closes (green stars). However, these conditions are not sufficient to create a cage with component (c), as there are two local minima in the orientation of the component (red stars).}
    \label{fig:fig_components}
\end{figure}

Fig. \ref{fig:limit} illustrates some slices between limit orientations where these conditions hold. While these conditions might seem restrictive, these can be used to represent a very large set of cages. We provide a geometric intuition of the different type of cages we can characterize in Fig. \ref{fig:fig_components}. We will derive a set of convex-combinatorial constraints to impose these conditions and yield an optimization formulation of the caging problem.

\subsection{Model Overview}

We make the following assumptions:
\begin{enumerate}
\item The object $\mathcal{O}$ is represented as the union of $M$ convex polygons, with a boundary that consists of $L$ facets.
\item The manipulator $\mathcal{M}$ is represented as a set of $N$ point-fingers.
\end{enumerate}
To include the conditions above, we discretize \co\ in $S$ $\mathcal{C}$-slices, similar to \cite{varava2017caging}, and impose that the manipulator bounds a component of free-space in each slice. Moreover, we make sure that the cage is not broken between slices by imposing continuity conditions between slices. \vspace{6pt}

We note that formulating this model requires continuous variables to represent the position of the manipulator, and binary variables to represent the discrete connectivity relation between $\mathcal{C}$-obstacles. Therefore, we introduce two sets of constraints to our model:
\begin{itemize}
\item \textbf{At each slice:} We require that a subset of the $\mathcal{C}-$obstacles forms a loop around the configuration of the object $[q_x,q_{y}]^T$. This ensures that there is a compact-connected component at each slice, as illustrated in Fig. \ref{fig:graph}. \vspace{6pt}
\item \textbf{For all orientations:} we constrain that either the object is caged for all $360^\circ$ or that $\mathcal{C}_{free}^{compact}(\mathcal{O})$ is bounded by two \textit{limit orientations}. Moreover, we require that the loop at each $\mathcal{C}-$slice maps continuously, without breaking, into the loop of its adjacent $\mathcal{C}-$slices. Through this, we can ensure that all components are connected between $\mathcal{C}-$slices, forming the component $\mathcal{C}_{free}^{compact}(\mathcal{O})$ that encloses $q$.
\end{itemize}
In sections IV and V, we will show that these conditions are sufficient to guarantee that $q$ is caged by $\mathcal{M}$. Furthermore, the use of convex-combinatorial constraints allows us to provide guarantees in terms of optimality and convergence.


\section{Constructing loops at each slice}
Here, we describe the set of conditions required to create a compact-connected component of free-space at one $\mathcal{C}-$slice. For notation convenience, we will refer to the  free-space of \cso\ as \csfree, and its enclosed component, where $q$ is, as $\mathcal{C}_{c}(\mathcal{O},\theta_s)$. To create $\mathcal{C}_{c}(\mathcal{O},\theta_s)$, we require that all the $\mathcal{C}-$obstacles in \cso\ form a loop. Since the object is decomposed into $M$ convex polygons, the problem of enforcing such a loop reduces to constructing a directed graph with the edges representing polygons intersection between $\mathcal{C}$-obstacles. Finally, we must also require that such loop encloses the configuration $[q_x,q_y]^T$ at the slice with fixed orientation $q_{\theta}$. \vspace{6pt}

\subsubsection{Existence of a loop} 
\begin{figure}[t]
		\centering
		\includegraphics[height=0.28\linewidth]{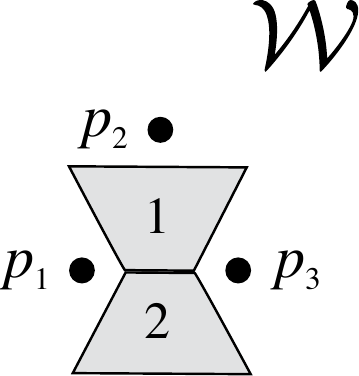} 
        \includegraphics[height=0.28\linewidth]{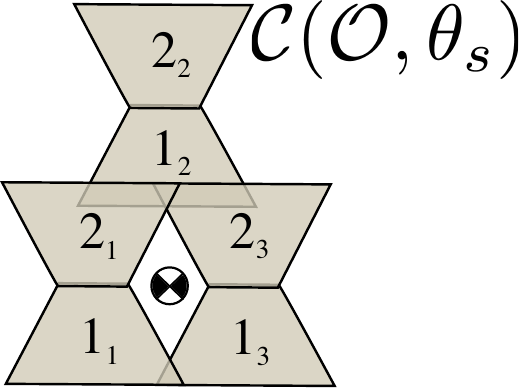} 
		\includegraphics[height=0.28\linewidth]{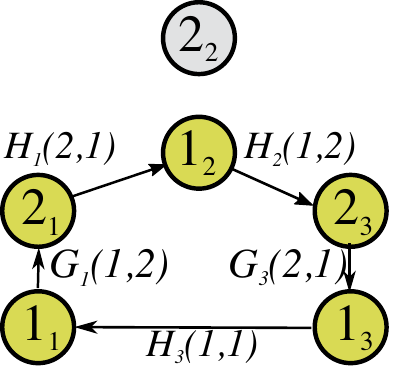}
		\caption{\small The caging loop for a slice with three fingers in $\mathcal{W}$ (left), \cso\ (center) and its corresponding connection graph (right)}
		\label{fig:graph}
        \vspace{-12pt}
\end{figure}

In order to compose an enclosing piecewise polygonal loop, we must determine which of the polygons on each $\mathcal{C}-$obstacles are part of the loop and their direction in the directed graph. To this end, we represent each polygon as a node, and add an edge between each pair of polygons that must intersect. Fig. \ref{fig:graph} illustrates this construction. For now, and for simplicity, we assume that all fingers must be part of this loop. \vspace{6pt}

For each finger, let us introduce a binary matrix $H_{n}\in\{0, 1\}^{M\times M}$, where $H_{n}(i, j) = 1$ if the $i_{th}$ polygon on $\mathcal{C}-$obstacle $n$ intersects with the $j_{th}$ polygon on $\mathcal{C}-$obstacle $n+1$. Mathematically if we denote the $i_{th}$ polygon in the $n_{th}$ finger as $\mathbf{P}_{i,n}$, then we impose the constraint:
\begin{equation}\label{eq:eq1}
		H_{n}(i, j) \Rightarrow \exists \text{ point } \mathbf{r}_n \in \mathbb{R}^2 \ \text{s.t.} \ \mathbf{r}_n \in \mathbf{P}_{i,n} \cap \mathbf{P}_{j,n+1}
\end{equation}
where we transcribe the $\Rightarrow$ (implies) operator as linear constraints through big-M formulation\footnote{For a binary $\mathbf{B}$, we have $\mathbf{B} \Rightarrow A x \leq b$ is equivalent to $A x \leq b + M(1-\mathbf{B})$ with $M$ being a large positive number. This allows us to represent conditionals within the optimization model through linear constraints.} \cite{bertsimas2005optimization}. Furthermore, to make each finger a part of the loop, we enforce the constraint:
\begin{equation}
	\sum_{i, j}H_{n}(i, j) = 1, \ \forall n \label{eq:all_fingers_in_loop}
\end{equation}
meaning that there is one and only one directed edge from $\mathcal{C}$-obstacle $n$ to $\mathcal{C}$-obstacle $n+1$. Since these connections are enforced for $n = 1, \hdots, N$, there is a directed loop of obstacles in \cso. In addition to using $H_n$ to represent the connectivity between two different C-obstacles, we introduce a matrix $G_n \in \{0,1\}^{M \times M}$ to denote if an edge in the $\mathcal{C}-$obstacle connection graph is in the loop. In the case that $G_n(i,j) = 1$ the graph has an edge going from polygon $i$ to polygon $j$ on the $n_{th}$ $\mathcal{C}-$obstacle. Then, the fingers create a closed loop if:
\begin{equation}
H_{n-1}(i,j) \Rightarrow \exists k, l \ \text{s.t.} \ G_n(j, k) + H_{n}(j,l) = 1 \label{eq:outbound_edge1}
\end{equation}
\begin{equation}
G_n(p, q) \Rightarrow \exists s,r \neq p \ \text{s.t.} \ G_n(q, r) + H_{n}(q, s) = 1\label{eq:outbound_edge2}
\end{equation}
\eqref{eq:outbound_edge1} and \eqref{eq:outbound_edge2} combined guarantee that for each node with an inbound edge, there is one and only one outbound edge, thus we have a loop in \cso. In the special case of a two-finger manipulator, since $H_{n,n+1}$ and $H_{n-1,n}$ have the same value, we need to further constrain that $l \neq i$ in \eqref{eq:outbound_edge1}.\vspace{6pt}


\subsubsection{Configuration enclosing} The previous constraints ensure the existence of a compact-connected component $\mathcal{C}_{c}(\mathcal{O},\theta_{s})$ in each slice. It is important to note that the enclosing loop implies the existence of  piecewise-polygonal curve that lives in the $\mathcal{C}-$obstacles. In our case, we define this curve by connecting points that live in the intersection of polygons in the loop. However, this does not guarantee that $q$ is contained in  $\mathcal{C}_{free}^{compact}(\mathcal{O})$. For this, we note that enclosing $q$ in $\mathcal{C}_{free}^{compact}(\mathcal{O})$ requires enclosing $[q_x,q_y]^T$ in $\mathcal{C}_{c}(\mathcal{O},q_{\theta})$. In order to incorporate this constraint to the model, we introduce Remark 1:\vspace{6pt}

\textbf{Remark 1 \cite{shimrat1962algorithm}:} \textit{If a linear ray that originates from a point $r$ has an odd number of intersections with a closed curve, then the point $r$ falls in the interior of such curve.}\vspace{6pt}

This remark is valid except for degenerate cases, when the ray is parallel to a segment of the curve or the area enclosed by the curve is a single point. However, such scenarios can be easily avoided in our analysis, as the enclosing curve is contained within the $\mathcal{C}-$obstacles. To incorporate this condition, we constrain the number of intersections between the ray and the line segments as a convex-combinatorial constraint. For this, we decompose the area that covers each possible line segment of the loop into 4 square regions, parallel to the ray, and introduce a binary decision matrix $F \in \{0,1\}^{N \times M \times 5}$, such that:
\begin{enumerate}
	\item $F(n,m,1)$ through $F(n,m,4)$ assign $[q_x,q_y]$ to one of four rectangular regions enclosing the line segment starting in the $m_{th}$ polygon of the $n_{th}$ finger.
    \item $F(n,m,5)$ is set to $1$ if the $m_{th}$ polygon of the $n_{th}$ finger is not part of the loop.
\end{enumerate}
Here, we assign $F(n,m,1)$ to the region that is parallel to the ray and below the line segment. Because of this, $F(n,m,1) = 1$ implies that the ray intersects the segment. A visualization of this can be seen in Fig. \ref{fig:region}. Then, we introduce the following pair of sufficient constraints:\vspace{-6pt}
\begin{equation}
    \begin{cases}
	\sum_{n,m} F(n,m,1) \ \text{is an odd number}\\
	\sum_{i = 1}^{5} F(n,m,i) = 1 \,\ \forall n,m 
	\end{cases}
\end{equation}
To transform (CT5) into a set of linear constraints, we introduce the following lemma:\vspace{6pt}

\begin{figure}[b]
	\vspace{-12pt}
	\centering
	\includegraphics[width=0.4\linewidth]{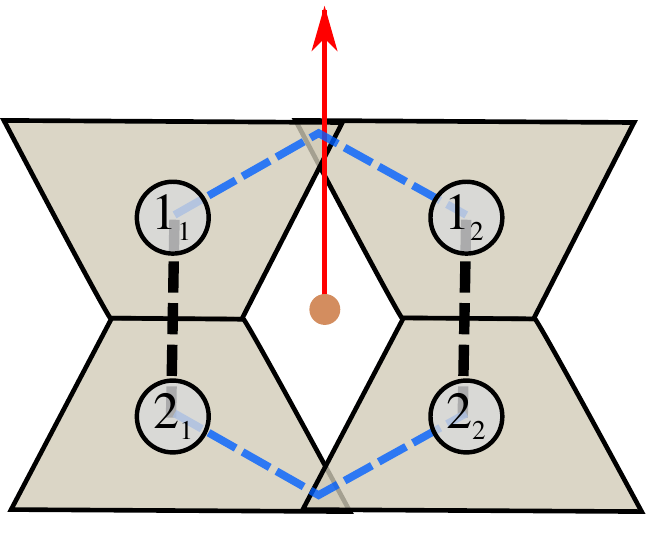} \hspace{12pt}
	\includegraphics[width=0.4\linewidth]{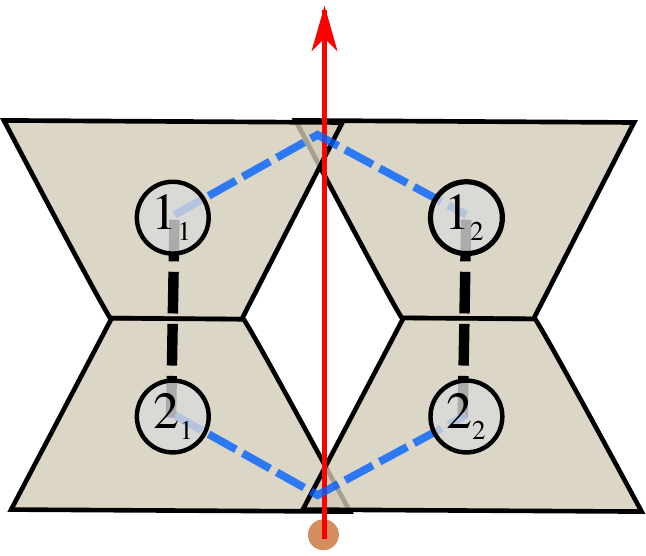}
	\caption{\small A point lies within a loop if a ray originating from that point has an odd number of intersections (blue) with the edges of the loop. Left: the point lies inside the polygon, and the ray has an odd number of intersections (one). Right: the point lies outside the polygon, and the ray has an even number of intersections (two).}
	\label{fig:ray}
\end{figure}

\textbf{Lemma 1.} \textit{The summation of binary variables $\sum_{i = 1}^{n} b_i$ is an odd number if and only if $b_1 \ XOR \ b_2 \ \dots \ XOR \ b_n  = 1 $}.\vspace{6pt}

Where the $XOR$ operator can be transcribed as linear constraints on the binary variables \cite{12118}. \vspace{6pt}

\begin{figure}[t]
		\centering
		\includegraphics[height=0.4\linewidth]{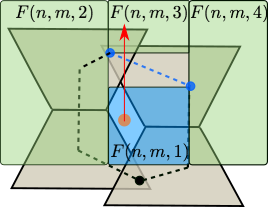}
		\caption{\small Region assignment of  $q$ (red dot) depending on the value of $F(n,m)$ and the direction of the linear ray (red arrow), for a line segment of the enclosing loop (blue). Note that, since the region is parallel to the ray, the ray always intersects the segment when $q$ is assigned with $F(n,m,1)$.}
		\label{fig:region}
        \vspace{-16pt}
\end{figure}

\subsubsection{Non-Penetration Constraints}

Additionally, we must prevent the fingers from penetrating the object. For this, we partition the 2D collision free workspace $\mathcal{W} \setminus\mathcal{O}$ into a set of $N_r$ convex regions, represented as:
	$$
	\mathcal{R}_i = \{ x \in \mathbb{R}^2 | A_i x \leq b_i  \}
	$$
and then constrain that each $p_n$ finger lies in one of these regions. To do this, we introduce a binary decision matrix $R \in \{0,1\}^{N_r \times N}$ such that:
\begin{eqnarray}
	R_{r,n} \Rightarrow A_i p_n \leq b_i \hspace{12pt} \text{and} \hspace{12pt} 
	\sum_{r = 1}^{N_r} R_{r,n} = 1, \forall n
\end{eqnarray}
Where we again transcribe the $\Rightarrow$ operator via big-M formulation. This ensures that each finger lies in only one of the regions. Note that this constraint also ensures that $q$ lies in the interior of ${\mathcal{C}}_{c}(\mathcal{O},\theta_{s})$ without penetrating any $\mathcal{C}-$obstacle. Sect. V discusses how these loops must interconnect in between slices to create a compact-connected component $\mathcal{C}_{free}^{compact}(\mathcal{O})$ in $\mathcal{C}$.

\begin{figure}[b]
		\centering
		\includegraphics[height=0.4\linewidth]{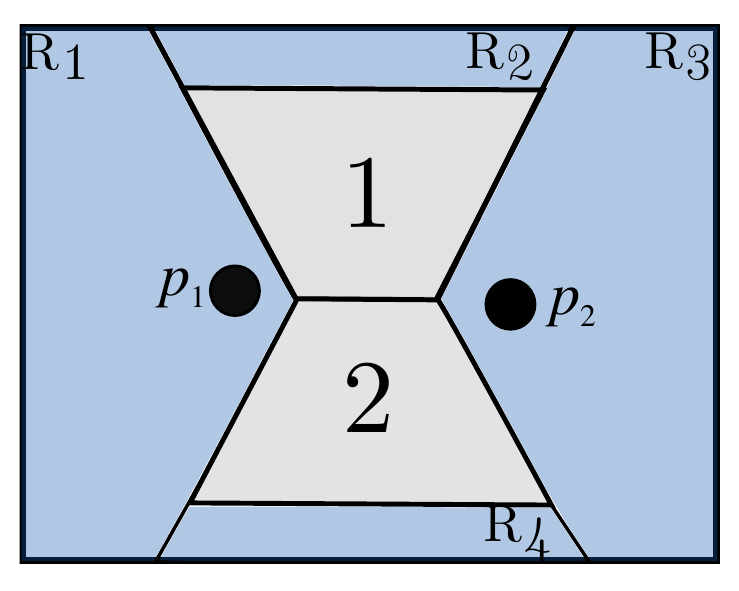}
		\caption{\small Collision-free region assignment of the fingers. In this example, the complement of the object is segmented in $N_r = 4$ polygonal regions $R_1,\dots,R_4$ where the fingers $p_1, p_2$ must lie. Note that a finger can lie in two different regions simultaneously, despite only being assigned to one.}
		\label{fig:region}
\end{figure}

\section{Constructing a cage from loops}

In order for an object to be caged, the compact-connected components formed at the $\mathcal{C}-$slices must be interconnected, such that there are no escaping paths in the space between slices. Furthermore, we always have that $\mathcal{C}_{free}^{compact}(\mathcal{O})$ either repeats periodically in the orientation dimension or closes at two points (limit orientations), in order to remain compact. In this section, we present a set of sufficient conditions that guarantee that the object configuration is caged in all $\mathcal{C}$.\vspace{6pt}

\subsubsection{Limit Orientations} 
\begin{figure}[t]
		\centering
\includegraphics[height=0.5\linewidth]{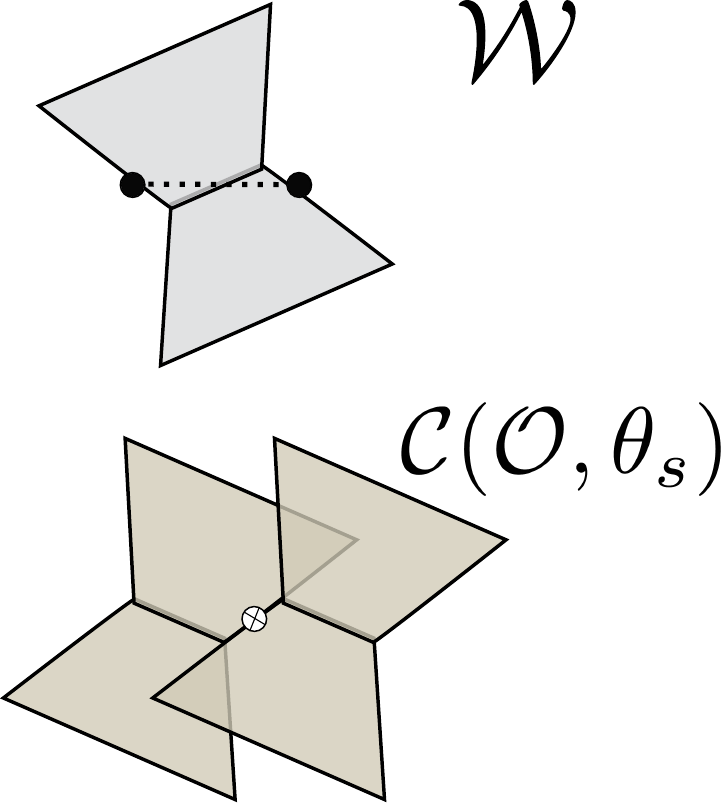}\hspace{12pt}
\includegraphics[height=0.5\linewidth]{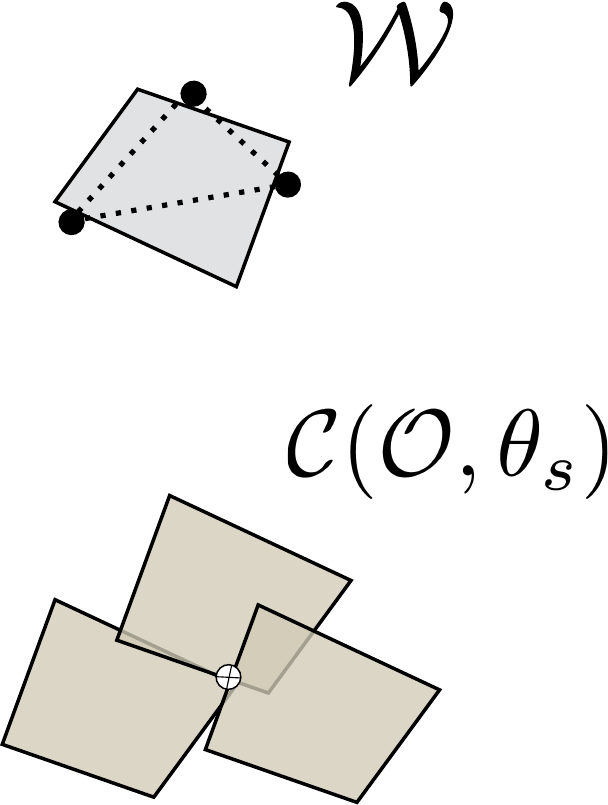}
		\caption{\small Two examples of limit orientations being reached for two fingers (left: two opposite facet) and three fingers (right: three non-parallel facets).}
		\label{fig:limitorient}
		\vspace{-12pt}
\end{figure}

For most objects of interest, caging in all 360$^\circ$ cannot be achieved, except when a manipulators with sufficiently many fingers fences the object. Hence, in order to create $\mathcal{C}_{free}^{compact}(\mathcal{O})$, the model must determine if there is a pair of limit orientations $\theta_l,\theta_u$. To account for this, we introduce a binary variable $\Theta \in \{0,1\}^{S}$, such that $\Theta_s = 0$ implies that the $s_{th}$ slice has not reached a limit orientation and, thus, the $\mathcal{C}-$obstacles must create a loop. Then, we introduce the following constraint:
\begin{equation}
	\begin{cases}
    1 - K \Theta_s \le \sum_{r = 1}^{N_r} R_{r,n}(\theta_s) \le 1 + K \Theta_s\\
	1 - K \Theta_s \le \sum_{i, j}H_{n}(i,j,\theta_{s}) \le 1 + K \Theta_s
	\end{cases}
\end{equation}
Where $K \in \mathbb{R}$ is a big number, this ensures that a loop is only enforced in the space between two limit orientations when these exist. We constrain that $\Theta_s = 1$ when the component of \cfree\ in the slice gets reduced to zero area. Moreover, assuming slices are ordered by increasing orientation, we add the condition that $\Theta_s = 1$ implies $\Theta_{s+1} = 1$ for positive orientations and $\Theta_{s-1} = 1$ for negative ones. Reaching this zero-area condition is dependent on the number of fingers and the $L$ facets of the object. As illustrated in Fig. \ref{fig:limitorient}, some scenarios when this zero-area condition is met are:\vspace{3pt}
\begin{enumerate}
	\item Two fingers: the fingers make contact with two parallel but opposite facets of the object, as in Fig. \ref{fig:limit}.\vspace{3pt}
	\item Three or more fingers: either three of the fingers are in contact with non-parallel facets or four of the fingers are in contact with non-co-directional facets.\vspace{3pt}
\end{enumerate}
Here, opposite refers to facets with parallel normals and opposite direction. It is important to note that, since the fingers create a loop that encloses $q$, \textit{concave vertices} in the object can be considered opposite and non-parallel to any facet with $180^\circ$ difference to the arc of the vertex.\vspace{6pt}
   
Finally, in order to determine when a limit orientation has been reached, we introduce a binary matrix $T_s \in \{0,1\}^{N \times L}$, at each slice, such that $T_{s}(n,f) = 1$ implies that the $n_{th}$ finger be in contact with facet $f$ at the $s_{th}$ slice, for some position of the object within the slice. Depending on the object and the number of fingers, the $T$ matrix will be constrained to determine when a limit orientation has been reached. Denoting $\mathcal{L}_\mathcal{O}$ as the set of facet assignments resulting in a limit orientation, we constrain:
\begin{equation}
T_s \in \mathcal{L}_\mathcal{O} \Rightarrow \Theta_s = 1
\end{equation}
this constraint guarantees that the model will detect if two limit orientations are necessary. Examples of some contact conditions required to reach a limit orientation can be seen in Fig. \ref{fig:limitorient}. As a reference, the idea of caging between two limit configurations can be seen as equivalent to finding a critical point of the inter-finger distance function in the contact space of the object \cite{allen2015two,bunis2018equilateral}. \vspace{6pt}

\subsubsection{Continuous Boundary Variation} The constraints described above ensure that there is a compact-connected component of free-space at each slice and that $q$ lies in one of them. However, escaping paths might still exist in the space between slices, where orientation changes along with the motion \cite{rodriguez2012caging}. To avoid this, we propose a set of sufficient conditions, which ensures that the union of all components $\mathcal{C}_{c}(\mathcal{O},\theta_{s})$ encloses $q$ in a compact-connected component of free-space. First, we define the notion of a boundary:\vspace{6pt}

\textbf{Definition 2} (Component Boundary): Given a component $C \in \mathcal{C}$, we define its boundary $\partial(C) \in \partial_C$ as the intersection of the closure of $C$ and the closure of its complement, where $\partial_C$ is the set of all boundaries.\vspace{6pt}

Let us define the map $f: \mathcal{C} \times \mathbb{R} \rightarrow \partial_C$, which parametrizes the boundary of a component $C$ with respect to a variable $\theta$. Next, we introduce the function $\delta_H: \partial_C \times \partial_C \rightarrow \mathbb{R}$ which returns the Hausdorff distance \cite{aspert2002mesh} between two boundaries $C_1$ and $C_2$. Finally, we say that the boundary of $C$ \textit{varies continuously} with respect to a variable $\theta$ if $ \underset{\Delta \theta \rightarrow 0}{\text{lim}} \ \delta_H(f(C,\theta),f(C,\theta+\Delta \theta)) = 0, \forall \theta$. Then, let us propose Theorem 1:\vspace{6pt}

\textbf{Theorem 1} (Continuous Boundary Variation). \textit{Given a pair of $\mathcal{C}-$slices with compact-connected components of free-space $C_{c}(\mathcal{O},\theta_{s}),C_{c}(\mathcal{O},\theta_{s+1})$, if the boundary $\partial({\mathcal{C}}_{c}(\mathcal{O},\theta_{s}))$ varies continuously w.r.t. orientation changes until $\partial( \mathcal{C}_{c}(\mathcal{O},\theta_{s+1}))$, then there will exist a compact-connected component in the free-space bounded by the slices, intersecting \cso\ and $\mathcal{C}(\theta_{s+1})$ in $C_{c}(\mathcal{O},\theta_{s})$ and $C_{c}(\mathcal{O},\theta_{s+1})$ respectively.}\vspace{6pt}
	
\begin{figure}[b]
	\centering
	\includegraphics[width=0.99\linewidth]{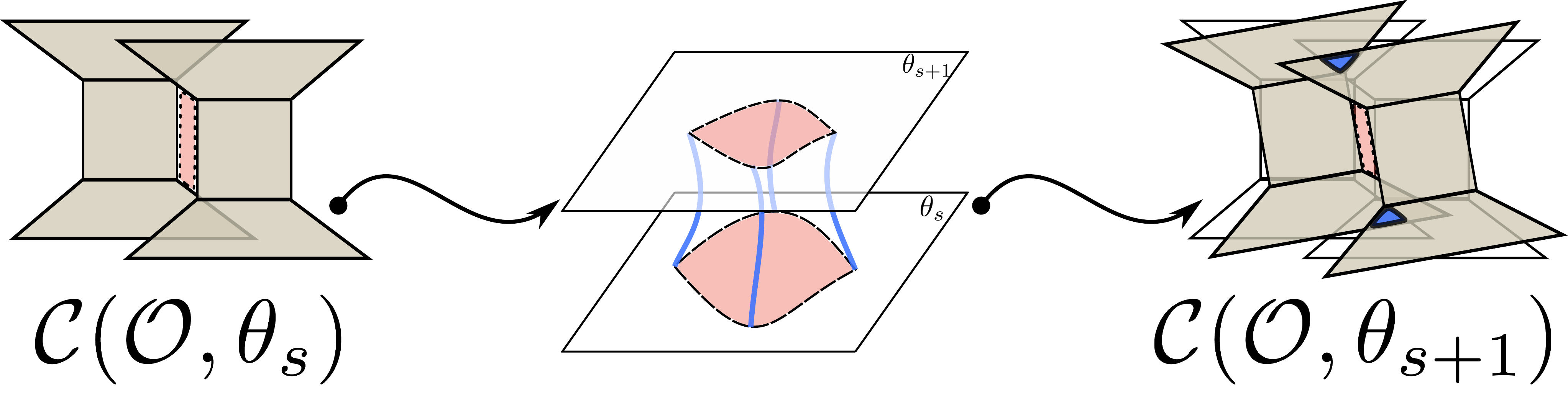}
	\caption{\small If the connecting polygons have a continuous intersection (blue) between adjacent slices, there is a continuous variation of the compact-connect component $\mathcal{C}_{c}(\mathcal{O},\theta)$ (red).}
	\label{fig:cbv}
\end{figure}

A proof of this theorem is sketched in Appendix A. By applying Theorem 1, we can guarantee that the loops defined at each slice form a compact-connected component of free-space, as long as their boundaries vary continuously between adjacent slices. To include this condition within the model, we introduce lemma 1:\vspace{6pt} 

\textbf{Lemma 1}. For a pair of $\mathcal{C}-$slices $\mathcal{C}(\theta_s), \mathcal{C}(\theta_{s+1})$ containing a closed loop of $\mathcal{C}-$obstacles, with intersections between polygons $\mathbf{P}_{i, n}$ and $\mathbf{P}_{j, n+1}$, if there exists a point (fixed to $\mathbf{P}_{j, n+1}$) that remains in each intersection $\mathbf{P}_{i, n} \bigcup \mathbf{P}_{j, n+1}$ during the rotation from slice $s$ to $s+1$, then there is a continuous boundary variation of the loop between $\mathcal{C}(\mathcal{O},s)$ and $\mathcal{C}(\mathcal{O},s+1)$. \vspace{6pt} 

A detailed proof of lemma 1 will be included in an extended version of this paper. Note that, since rotations map the $\mathcal{C}-$obstacles continuously along the $\theta$ axis, this condition forbids the component between each pair of slices to \textbf{discontinuously} separate in several {disconnected} components or become unbounded.

As consequence of lemma 1, A simple convex constraint to ensure continuous boundary variation, is to require the same discrete connections between polygons in adjacent slices, as shown in Fig. \ref{fig:cbv}. To model this constraint, for each pair of $\mathcal{C}-obstacles$ we denote the intersecting point as $x$, and introduce its position fixed to $\mathbf{P}_{j, n+1}$ as new decision variable $\leftidx{^{j, n+1}}x$. Here, the left superscript $^{j, n+1}$ denotes the position is expressed in the local coordinate of $\mathbf{P}_{j, n+1}$. We first impose the constraint:
\begin{align}
    \leftidx{^{j, n+1}} x \in \mathbf{P}_{j, n+1}\label{eq:point_fixed_in_polygon}
\end{align}
such that this point is always in  $\mathbf{P}_{j, n+1}$. On the other hand, to constrain that point $x$ remains in $\mathbf{P}_{i, n}$ during rotation $\theta\in[\theta_s,\theta_{s+1}]$, we consider the trajectory of $x$ observed in $\mathbf{P}_{i, n}$'s coordinate as a function of $\theta$:
\begin{align}
    \leftidx{^{i, n}}x(\theta) = \leftidx{^{j, n+1}}x + R(\theta)^T(c_{j, n+1} - c_{i, n})\nonumber
\end{align}

\begin{figure}[t]
		\centering
		\begin{subfigure}{0.25\textwidth}
  \includegraphics[width=0.9\textwidth]{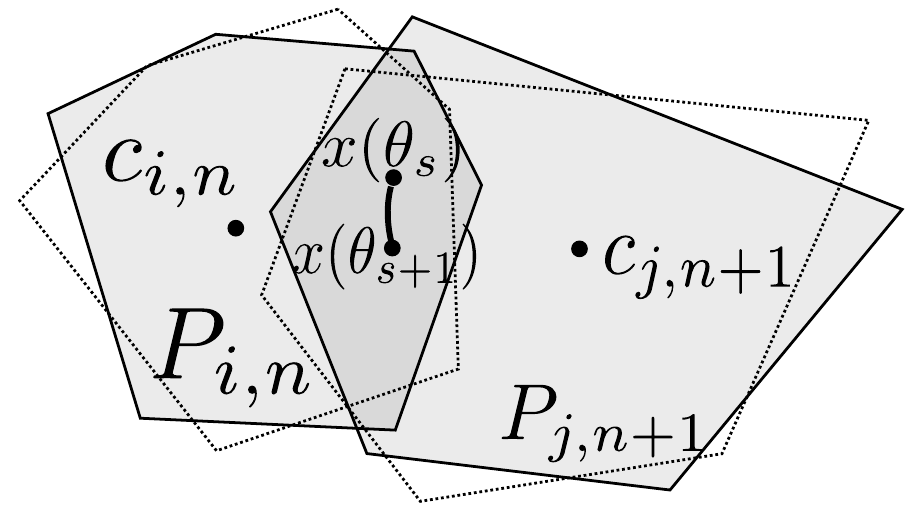}
        \subcaption{\small The polytopes rotate to the light colored orientation. The trajectory of the intersecting point $x$ is an arc.}
        \label{fig:ins1}
    \end{subfigure}
    \hspace{1pt}
    \begin{subfigure}{0.22\textwidth}
        \includegraphics[width=0.7\textwidth]{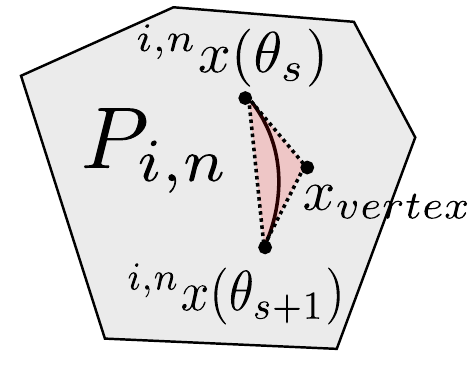}
        \subcaption{\small Arc observed from $P_{i, n}$'s coordinate. The arc is enveloped by a triangle.}
        \label{fig:ins2}
    \end{subfigure}
		\caption{\small  In order to avoid that the loop breaks between slices, we require the polygons always intersect at a point $x$ during rotation.}
		\label{fig:ins}
		
\end{figure}

where $R(\theta)\in\mathbb{R}^{2\times 2}$ is the rotation matrix for angle $\theta$, and $c_{i, n}, c_{j, n+1}$ are the center of rotation for $\mathbf{P}_{i, n}$ and $\mathbf{P}_{j, n+1}$ respectively. Hence, $\leftidx{^{i, n}}x(\theta)$ is an arc on a circle, with $\leftidx{^{j, n+1}}x$ being the center of the circle, and $|c_{j, n+1} - c_{i, n}|$ the circle radius drawn in Fig.\ref{fig:ins}. Constraining point $x$ to stay in polygon $\mathbf{P}_{i, n}$ is equivalent to requiring this arc being in $\mathbf{P}_{i, n}$. Since the arc lies within the triangle formed by the two ends of the arc ($\leftidx{^{i, n}}x(\theta_s), \leftidx{^{i, n}}x(\theta_{s+1})$) and the intersecting point $x_{v}$ between two arc tangents at the two ends, a sufficient condition of arc in the polygon is that all three vertices of this triangle are in the polygon. The position of $x_{v}$ can be computed as:
\begin{align}
    &x_{v} =\label{eq:arc_in_polygon}\\ &\leftidx{^{i, n}}x(\theta_s) + \left(\mathbf{I}_{2\times2} + R(-90^\circ)\tan(\frac{\theta_{s+1} - \theta_s}{2})\right)(c_{j, n+1} - c_{i, n}) \nonumber
\end{align}
again as a linear function of our decision variables. We activate the constraints above when the two polygons are intersecting, as:
\begin{equation}
H_{n}(i,j,\theta_s) \Rightarrow \leftidx{^{i, n}}x(\theta_s), \leftidx{^{i, n}}x({\theta_{s+1}}), x_{}\in \mathbf{P}_{i, n}
\end{equation}
Through this constraint, we ensure that the intersecting polygons of \cso\ remain connected for all orientations between $\mathcal{C}(\theta_s)$ and $\mathcal{C}(\theta_{s+1})$. Hence, the free-space component boundary only expands or contracts continuously when rotating between slices. Because of this, there is a single compact-connected component of free-space that enclosed the configuration of the object $q$. 

\section{Implementation and Results}

In this section we implement an optimization-based cage-finding algorithm, derived from the proposed model. We also validate the tractability and versatility of this formulation by synthesizing cages for different planar geometries, under different sets of constraints.

\subsection{Formulating Caging as Optimization}

Given an object segmented in $M$ polygons, a manipulator with $N$ fingers and sampling \co\ in $S$ $\mathcal{C}-$slices, we formulate the cage-finding algorithm as the feasibility problem $\mathbf{MIP1}$.

\begin{equation}\nonumber
\mathbf{MIP1:} \ \underset{\Theta, H, G, R, T, p}{\text{\text{find}}} \ \ \ \ \ \ p_1, \dots, p_N
\end{equation}
subject to:\vspace{6pt}
\begin{enumerate}
	\item  For all $S$ slices:
    \begin{itemize}
    	\item Existence of a loop (CT1)---(CT4).
        \item Non-penetration (CT6).
        \item Limit orientation constraints (CT7)---(CT8).
    \end{itemize}
   \item For slice containing $\theta_s = q_\theta$:
   \begin{itemize}
   \item Configuration enclosing (CT5).
   \end{itemize}
   \item Continuous Boundary Variation (CT9)---(CT11).\vspace{6pt}
\end{enumerate} 

Through this formulation, we apply our model to find cages on planar objects, with an arbitrary number of fingers.\vspace{12pt}

\subsubsection{Properties of the Model} The formulation of the problem as a Mixed-Integer Convex Program \cite{bertsimas2005optimization} provides several useful properties and guarantees. First and foremost, given sufficient time, a solver can always find the global solution to this type of optimization problem, providing a convergence guarantee. Secondly, if a convex cost function is added to \textbf{MIP1} the solver will always converge to its global optima (with worst-case exponential complexity). Finally, this formulation is versatile, as additional mixed-integer convex constraints can be added to the model \cite{dai2017global} without losing its properties. Because of this, if there is a cage that satisfies the conditions presented on this paper, the optimization problem will always find it. Similarly, if there is no cage that satisfies the sufficient conditions we define, the solver will always report so. 

\subsection{Model Validation}

\begin{figure}[h]
	\centering
	\includegraphics[width=\linewidth]{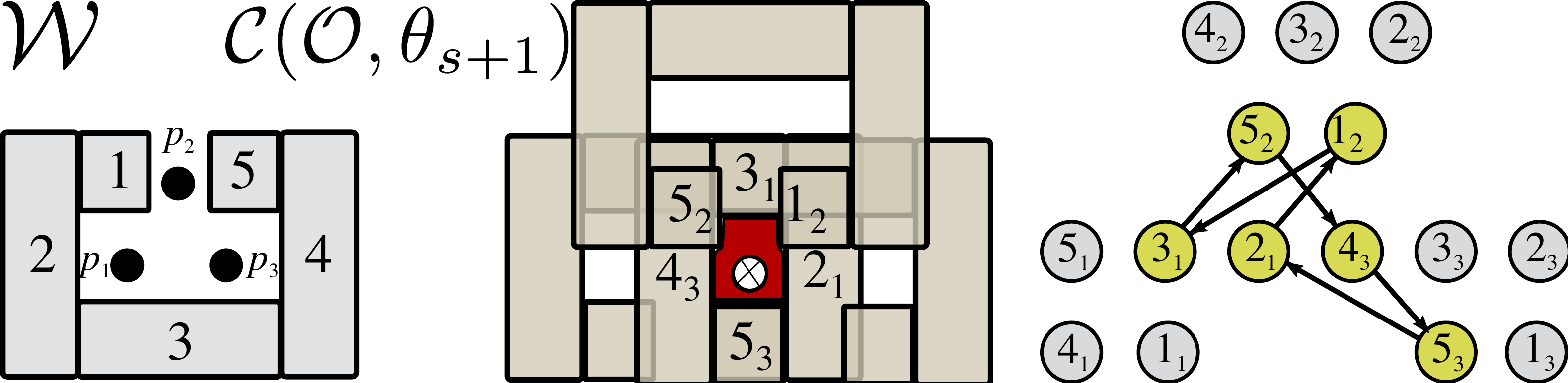}
	\caption{\small Example cage found with $\mathbf{MIP1}$ in the workspace (left), configuration space slice (center) and its connection graph (right).}
	\label{fig:res3}
\end{figure}

In order to test the proposed cage-finding formulation, we transcribe the optimization problem and solve it using off-the-self optimization software. All the tests are performed in MATLAB R2018b, on a Intel Core i9 laptop running Mac OS X High Sierra. We use Gurobi 8.0.0 \cite{gurobi} as our MIP solver. For all tests, we set the parameter $S$ to 9 slices, evenly distributed in a range between $-90^\circ$ and $90^\circ$. As a proof of concept, Fig. \ref{fig:res3} shows an example of a cage found with this approach, on a non-intuitive object where increasing finger dispersion does not guarantee a cage \cite{rodriguez2012caging}.

To illustrate the capabilities of the model, we will perform a set of tests showcasing its generality and versatility to include additional constraints.\vspace{6pt}

\begin{figure*}
    \centering
    \includegraphics[width=\linewidth]{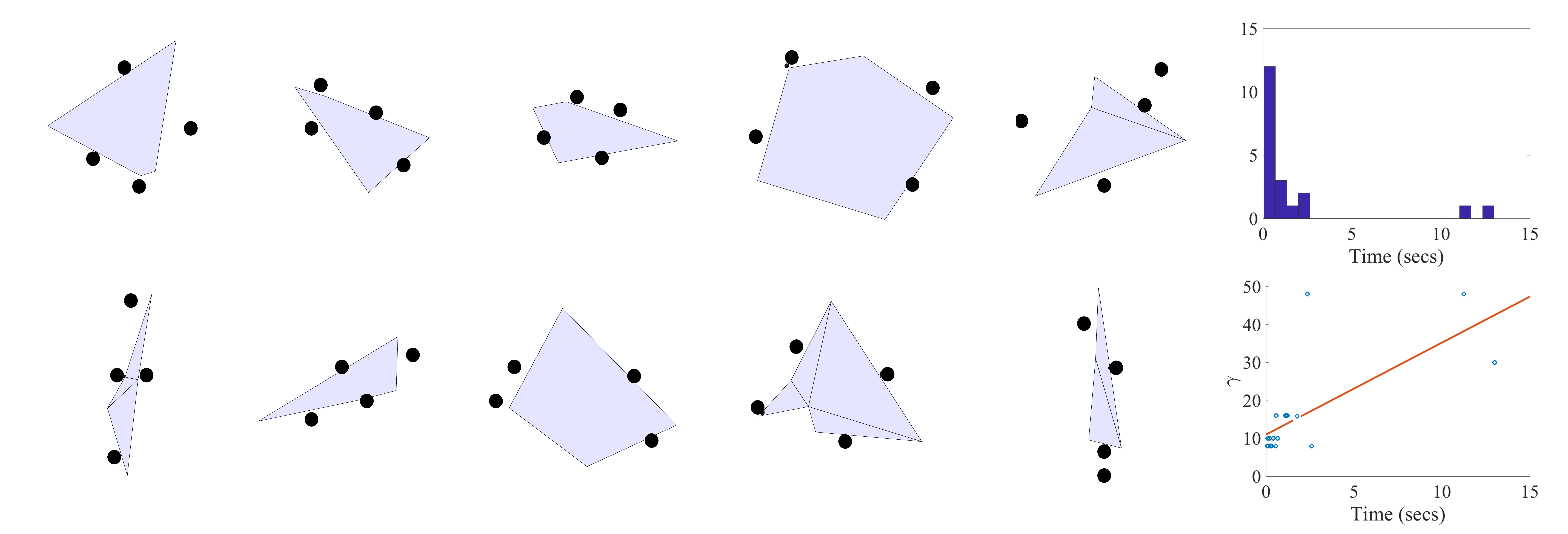}
    \caption{\small Synthesis of cage for a set of random polygonal objects. Our model is able to generalize across convexity and symmetry. Plots show a performance analysis of the model: computation time histogram (top) and time vs complexity index $\gamma$ (bottom).}
    \label{fig:res_rand}
    \vspace{-12pt}
\end{figure*}


\subsubsection{Caging random polygons}

To showcase the generality of this approach, we generate a set of 20 random polygonal shapes -- segmented in convex polygons through Delaunay triangulation \cite{fortune1995voronoi}. Afterwards, we call on the optimizer to find a cage for a manipulator with four fingers. We report the obtained cages for 10 of these shapes in Fig. \ref{fig:res_rand}, along with a brief performance analysis of the optimization problem. Our model can efficiently find the location of the fingers in order to cage each object regardless of its shape, handling non-convex and non-symmetric shapes. However, we note that caging with 2 and 3 fingers usually reports infeasibility without the appropriate slicing, as limit orientations might be hard to find in such cases. To deal with this limitation, we sliced the orientation component by taking into account the shape of the object and the angular separation between facets. When caging with four or more fingers, a cage is easily found with uniform slicing. Also, we assess the scalability of our model by computing a complexity index $\gamma = {M \cdot L \cdot R}$ ($\#$ of polygons $\times$ $\#$ of line segments $\times$ $\#$ of collision-free regions) for each object. Fig. \ref{fig:res_rand} shows a histogram of computation time (t) and performance indexes. Our results show that computation can be generally within 1-2 seconds, and scales well with shape complexity. \vspace{6pt}

\subsubsection{Caging with constraints} 

The optimization nature of this approach allows us to include additional constraints in the caging problem. For this, we study two cases of interest: (1) Caging with kinematic constraints, (2) Caging with two fingers against a fixed-environment (wall). In particular, we show cages found using two hands with parallel grippers with limited opening and cages using two unconstrained fingers and a wall. 

\vspace{6pt}

To model the kinematics of (1) we use the dimensions of an ABB YuMi$^{\tiny{\textregistered}}$ robot over a plane. These kinematics require that each pair of fingers maintains the same vertical position, while respecting the reachability of the hand (maximum and minimum separation between fingers). To incorporate the wall of (2), we distribute 5 fixed point-fingers above the object to be caged and allow two robot fingers to move freely over space. Other methods could also be used to model the wall, such as computing the $\mathcal{C}-$obstacle of a line segment directly to include it as part of the loop.

\vspace{6pt}

\begin{figure*}
    \centering
    \includegraphics[width=\linewidth]{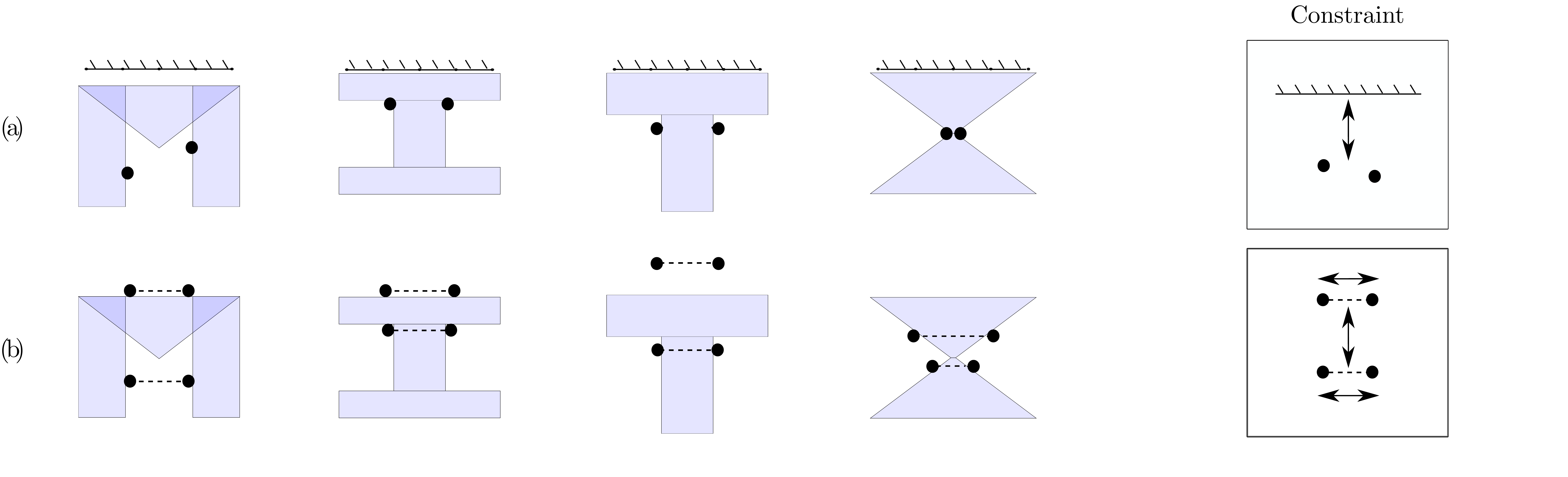}
    \vspace{-12pt}
    \caption{\small Synthesis of cages with additional constraints: (a) Cage using two fingers and a fixed environment, a wall segment located over the object (b) Caging with four fingers subject to the kinematic constraints of two parallel grippers.}
    \label{fig:res_const}
    \vspace{-12pt}
\end{figure*}

Under these conditions, we optimize cages for 4 different planar objects, shown in Fig. \ref{fig:res_const}. In contrast to traditional approaches, where these conditions would be derived analytically, we need to only include additional constraints that describe the manipulator $\mathcal{M}$ into the model. We also note that convergence to a cage is significantly faster in these scenarios, despite having more constraints in the model. The reason for this speed-up is that additional constraints reduce the search space by quickly discarding infeasible branches.

\section{Discussion}

In this paper, we have presented a novel convex-combinatorial model for planar caging, able to reason over arbitrary planar polygonal shapes with an arbitrary number of point fingers. The formulation is based on a set of sufficient conditions which can be transcribed as constraints within an optimization problem. To the best of our knowledge, this is the first optimization-based approach to formulate the caging condition. A key contribution of the work is the potential of the formulation to be compatible with other task constraints. For example: to include kinematics of the manipulator \cite{dai2017global} (e.g. coupled grippers), environment use \cite{Eppner15Environment} (e.g. caging with a wall), and reaching motions \cite{rodriguez2012caging} (e.g. cage $\rightarrow$ reach $\rightarrow$ grasp). Furthermore, we have shown how to implement a cage-synthesis algorithm derived from our model, easily solvable as an MIP. Our results showcase the versatility and convergence properties of this approach.

\vspace{6pt}

\paragraph{Limitations of the model}

The properties of this model come at the expense of limitations on the cages that are characterized by the constraints. First, the dependence on slicing makes the model sensitive to the selection of orientations used for finding a cage, particularly when shapes are non-symmetric. Second, the enclosing constraints (Section IV.2) are often restrictive, since $[q_x,q_y]^T$ might not be able to lie in any of the 4 regions covering a line segment, making (CT5) infeasible (e.g. caging a triangle with 3 fingers) . Finally, this model assumes all fingers are part of the enclosing loop and does not emphasize cages with a minimal number of fingers \cite{pereira2004decentralized}.

\vspace{6pt}

\paragraph{Future Work}

Future efforts should aim to make the model more flexible. First, by studying how object shape information can be leveraged for slice selection. Similarly, it is important to explore methods to reduce the complexity of the model, currently exponential in the worst case. This can potentially be done through the introduction of stronger conditions that reduce the combinatorial elements in the formulation. Furthermore, while the conditions in this paper are sufficient to guarantee a cage, in particular those derived from Theorem 1, we suspect that these could also become necessary and sufficient through a dense enough slicing. Hence, this potential \textit{``resolution completeness"} property of the model deserves further study. 

\vspace{6pt}

\paragraph{Source Code}

The entire source code used as part of this work is publicly available on GitHub: \href{https://github.com/baceituno}{https://github.com/baceituno}

\section*{Acknowledgments}

We would like to thank Anastasiia Varava, Jos\'e Camacho, Jos\'e Ballester, Sean Curtis and members of the MCube Lab for insightful discussions and advice during the development of this project.

\bibliographystyle{./IEEEtran}
\bibliography{./IEEEabrv,./references}

\appendix

\begin{figure}[b]
	\centering
	\vspace{-12pt}
	\includegraphics[width=0.9\linewidth]{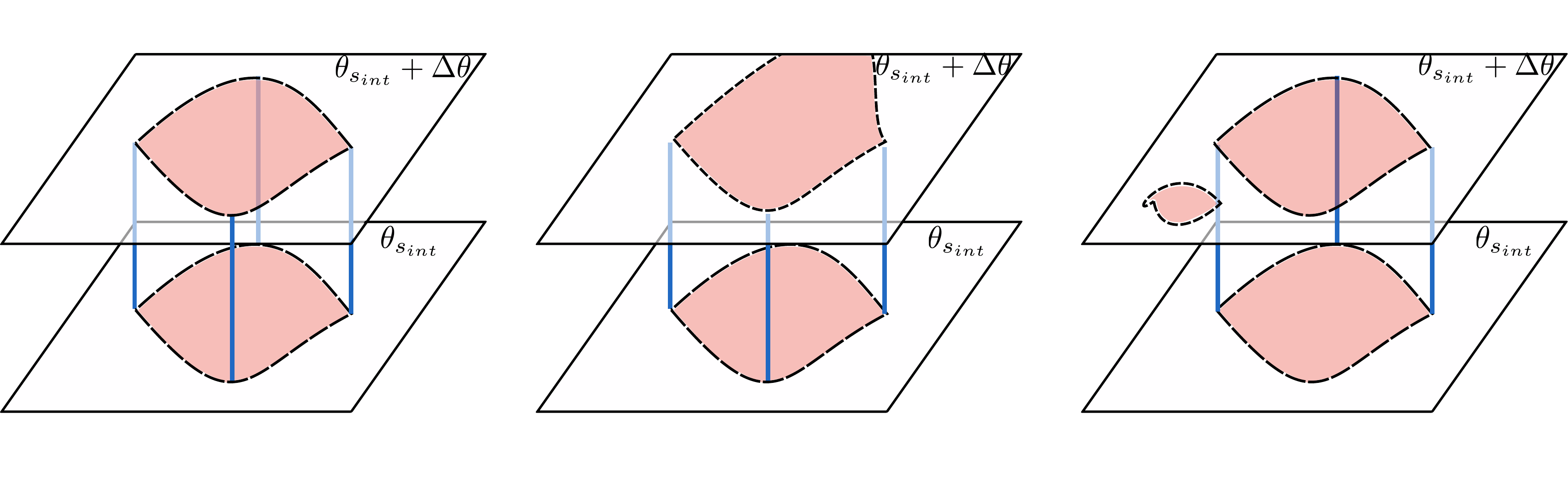}
	\caption{\small	Cases with -- (a) -- and without -- (b) and (c) -- continuous boundary variation used in the proof of theorem 1. All the cases where the component becomes unbounded (b) or disconnected (c) lose continuous boundary variation.}
	\label{fig:cbvnx}
\end{figure}

\textit{Proof Outline for Theorem 1.} The proof of this theorem follows the contradiction. Suppose that the component of free-space between the slices, which can be defined as the union $\mathcal{K} = \underset{\theta \in [\theta_s, \theta_{s+1}]}{\bigcup} C_{c}(\mathcal{O},\theta)$, has continuous boundary variation with respect to the orientation component, but is not compact or connected. Then if we analyze the boundary: 
\begin{enumerate}
    \item There must exist at least a slice in $\mathcal{K}$, such that there is no compact-component component of free-space.
    \item The component must either not be compact or transforms into several disconnected components after at least some slice $\mathcal{C}(\theta_{s_{int}})$.
\end{enumerate}
Therefore, we have at least one of the following scenarios: 
\begin{enumerate}
    \item The component of free-space becomes non-compact at $\mathcal{C}(\theta_{s_{int}})$, which implies that $\underset{\Delta \theta \rightarrow 0}{\text{lim}} \ \delta_H(\partial C_{c}(\mathcal{O},\theta_{s_{int}}),\partial  C_{c}(\mathcal{O},\theta_{s_{int}} + \Delta \theta)) = \infty$.
    \item At least a disconnected component of free-space appears at $\mathcal{C}(\theta_{s_{int}})$, which implies $\underset{\Delta \theta \rightarrow 0}{\text{lim}} \ \delta_H(\partial  C_{c}(\mathcal{O},\theta_{s_{int}}),\partial C_{c}(\mathcal{O},\theta_{s_{int}} + \Delta \theta)) > 0$.
\end{enumerate}
An illustration of these cases is shown in Fig. \ref{fig:cbvnx}. By hypothesis, $\underset{\Delta \theta \rightarrow 0}{\text{lim}} \ \delta_H(\partial C_{c}(\mathcal{O},\theta),\partial  C_{c}(\mathcal{O},\theta + \Delta \theta)) = 0, \forall \theta \in [\theta_s,\theta_{s+1}]$. Therefore, the non-existence of a compact-connected component results in a contradiction, concluding the proof. $ \blacksquare $.

\end{document}